\documentclass[sigconf]{aamas} 
\usepackage{balance} % for balancing columns on the final page

\usepackage[caption=false]{subfig}

\setcopyright{ifaamas}
\acmConference[AAMAS '22]{Proc.\@ of the 21st International Conference
on Autonomous Agents and Multiagent Systems (AAMAS 2022)}{May 9--13, 2022}
{Online}{P.~Faliszewski, V.~Mascardi, C.~Pelachaud,
M.E.~Taylor (eds.)}
\copyrightyear{2022}
\acmYear{2022}
\acmDOI{}
\acmPrice{}
\acmISBN{}

\acmSubmissionID{61}

\title[]{Empirical Estimates on Hand Manipulation are Recoverable: 
\\A Step Towards Individualized and Explainable Robotic Support in Everyday Activities}

\author{Alexander Wich}
\affiliation{
  \institution{Institute for Artificial Intelligence, University of Bremen}
  \city{Bremen}
  \country{Germany}}
\email{awich@uni-bremen.de}

\author{Holger Schultheis}
\affiliation{
  \institution{Institute for Artificial Intelligence, University of Bremen}
  \city{Bremen}
  \country{Germany}}
\email{schulth@uni-bremen.de}

\author{Michael Beetz}
\affiliation{
  \institution{Institute for Artificial Intelligence, University of Bremen}
  \city{Bremen}
  \country{Germany}}
\email{beetz@cs.uni-bremen.de}

\begin{abstract}

A key challenge for robotic systems is to figure out the behavior of another agent. The capability to draw correct inferences is crucial to derive human behavior from examples.

Processing correct inferences is especially challenging when (confounding) factors are not controlled experimentally (observational evidence). 
For this reason, robots that rely on inferences that are correlational risk a biased interpretation of the evidence.

We propose equipping robots with the necessary tools to conduct observational studies on people. 
Specifically, we propose and explore the feasibility of structural causal models with non-parametric estimators to derive empirical estimates on hand behavior in the context of object manipulation in a virtual kitchen scenario. 
In particular, we focus on inferences under (the weaker) conditions of partial confounding (the model covering only some factors) and confront estimators with hundreds of samples instead of the typical order of thousands. 
Studying these conditions explores the boundaries of the approach and its viability.

Despite the challenging conditions, the estimates inferred from the validation data are correct. Moreover, these estimates are stable against three refutation strategies where four estimators are in agreement. 
Furthermore, the causal quantity for two individuals reveals the sensibility of the approach to detect positive and negative effects. 

The validity, stability and explainability of the approach are encouraging and serve as the foundation for further research.

\end{abstract}


\newcommand{\BibTeX}{\rm B\kern-.05em{\sc i\kern-.025em b}\kern-.08em\TeX}

\begin{document}

%%% The following commands remove the headers in your paper. For final 
%%% papers, these will be inserted during the pagination process.

\pagestyle{fancy}
\fancyhead{}

%%% The next command prints the information defined in the preamble.

\maketitle 

%%%%%%%%%%%%%%%%%%%%%%%%%%%%%%%%%%%%%%%%%%%%%%%%%%%%%%%%%%%%%%%%%%%%%%%%

\section{Introduction}

The potential benefits for robots observing human behaviors are twofold. Having robots adopt people's know-how has the potential to incorporate key knowledge for robust control systems in workspaces that are made by and for human beings. In turn, robots modeling human behavior are in a better position to anticipate a person's intention, thereby improving assistance in joint tasks. In the same manner, robots adopting human behaviors act predictably for their human counterpart. 

At the age of two toddlers already display capabilities of reasoning over cause and effect \cite{Gopnik2010}. 
Inferring the cause of effects is a common capability in children, which enables them to understand their surroundings by simple observation, including the actions of other social agents \cite{journals/cogsci/SobelTG04}. % as done by preschoolers.
Causal reasoning enables children to observe the actions of other social agents and learn causal relationships from probabilistic events \cite{Waismeyer2014}. 

\begin{figure}
\centering
\includegraphics[width=0.9\linewidth,page=3]{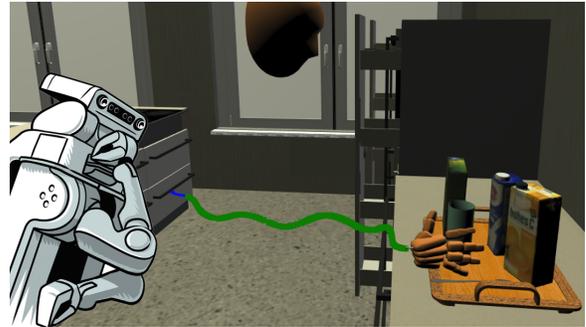}
\caption[Robots observing people]{
A robot observes a person set the table in a virtual kitchen scenario:
Why does this person sometimes choose the left hand over the right?
Does the proximity of the grasp play a role in hand-selection? 
And if so, why?
}\label{fig:problem_1a}
\Description[Robots observing people]{
A robot observes a person set the table in a virtual kitchen scenario:
Why does this person sometimes choose the left hand over the right?
Does the proximity of the grasp play a role in hand-selection? 
And if so, why?
}
\end{figure}

We envision robots able to acquire knowledge simply by observing their surrounding as illustrated in Figure~\ref{fig:problem_1a}. To enable observational capabilities in robots that resemble those of children, requires mechanizing inferences that are robust against bias. 

Ignoring confounding bias (a.k.a. reversals) can invalidate inferences in any domain \cite{Simpson51}. 
For instance, reversals affect a random uniformly distributed $2\times2\times2$ table with a probability of $1/60$ \cite{Pavlides:2009:HLS}. The probability that reversals would occur at random in path models involving two predictors and one criterion variable is approximately ten percent \cite{journals/ijec/Kock15}. 

\begin{figure}
\centering
%\hfill
\subfloat[\label{fig:problem_1b}Positive effect]{%
\includegraphics[width=0.5\linewidth]{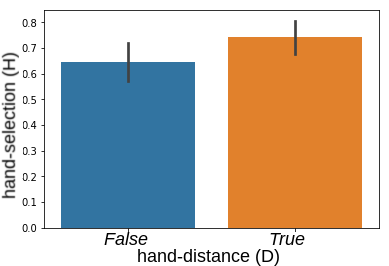}}
\subfloat[\label{fig:problem_1c}Negative effect]{%
\includegraphics[width=0.5\linewidth]{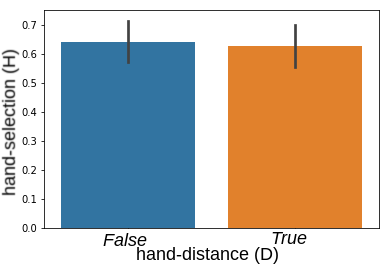}}
\caption[Bias in hand manipulation]{
Given identical data, Figure \ref{fig:problem_1b} and \ref{fig:problem_1c} show opposing conclusions when segregating and aggregating trends. This confirms the existence of reversal in our data and motivates the effort to account for bias. This data is further described in Table~\ref{tab:segmenteddataset}. 
}
\label{fig:problem}
\Description[Bias in hand manipulation]{Bias in hand manipulation motivates the efforts to take measures against it. Given identical data, Figure \ref{fig:problem_1b} and \ref{fig:problem_1c} show opposing conclusions when segregating and aggregating trends.}
\end{figure}

The data considered in this work is no exception to this. 
Figure~\ref{fig:problem} shows two opposing conclusions that are drawn on identical data. 
The disagreement of segregated and aggregated trends confirms the existence of reversal in the data considered here. 
Confounding can be confronted with a model as suggested in \cite{Pearl2014} as approached here. 
Section~\ref{sec:phase2} further describes the data.

The main proposition of this paper is for the robots to, literally, conduct observational studies.
We propose causal inference to equip robotic agents with the necessary observational skills.
The central question raised in this article is whether robots equipped with capabilities to perform observational studies extract reliably and transparently information from complex behavioral examples.

To this end, we take on the task to recover a non-trivial relationship on a few examples of hand-manipulation originating from uncontrolled sources of evidence. 
This is essentially the challenge children overcome when looking over to role models.

The steps required to equip robots with the necessary observational skills are introduced in Section~\ref{sec:phase1} to \ref{sec:phase3}. 
Subsequently, the validity of inferences are verified (Section~\ref{sec:validation}) and the stability of effects tested (Section~\ref{sec:robustness}). Then, inferences target evidence from two individuals whom are not influenced by any experimental design (Section~\ref{sec:moreeval}).
Last, individual insights on hand behavior are then presented (Section~\ref{sec:interpretation}). 

\section{Drawing Causal Conclusions on Observational Evidence}\label{sec:approach}

One of the necessary capabilities to derive behavior from evidence is for robots to draw causal conclusions. Because causal inferences mitigate bias, conclusions are closer to the truth than the correlational counterpart.

Drawing causal conclusions on uncontrolled evidence involves three steps. Step 1 defines a model for hand manipulation (described in Section \ref{sec:phase1}), step 2 introduces the evidence on which inferences are drawn (Section \ref{sec:phase2}), and step 3 formalizes causal reasoning (Section \ref{sec:phase3}). 

\subsection{Step 1: Defining the Model}\label{sec:phase1}

\begin{figure}[htbp] %[ht]
\centering
\includegraphics[width=\linewidth]{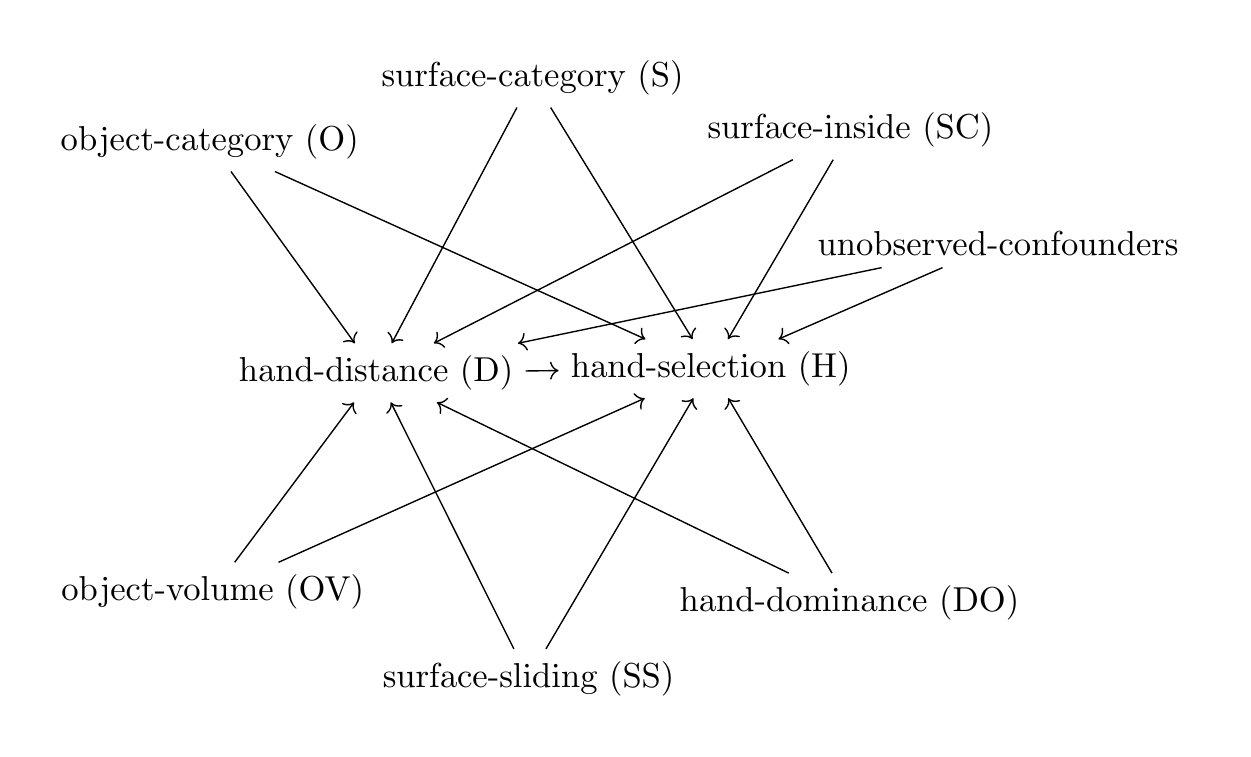}
\caption[Causal graph modeling hand behavior]{A causal graph modeling reach-to-grasp actions including two hand decisions along other relevant factors.
The validation hypothesis of this work corresponds to the arrow directed from hand-distance to hand-selection. 
}
\label{fig:graphical_model}
\Description[Causal graph modeling hand behavior]{A causal graph modeling reach-to-grasp actions including two hand decisions along other relevant factors. The validation hypothesis of this work corresponds to the arrow directed from hand-distance to hand-selection.}
\end{figure}

A graph encodes the model of hand manipulation as shown in Figure~\ref{fig:graphical_model}. In this diagram the nodes represent potential factors that affect hand manipulation, while arrows indicate the direction these factors affect each other.
For example, the arrow pointing from hand-distance to hand-selection states that the former affects the latter. 
Here, the choice of hand-selection refers to the hand, either left or right, a person grasps an object with. Similarly, the choice of hand-distance, either close or far from their body, refers to the straight-line distance between the head and hand at the time the person triggered a grasp. 

The arrows included in the causal graph are described as follows:
\begin{enumerate}
	\item \textbf{Hand-distance on hand-selection}: Hand selection occurs according to object proximity. For example, when targeting an object for which the left hand is closer than the right, the left hand is chosen.
	This behavior is supported by the kinesthetic hypothesis \cite{Gabbard2000,Gabbard2004}.

	% [Human Concepts]
	\item \textbf{Object on hand-distance}: The category of an object influences the distance a person would stretch their hand relative to their body when grasping objects. For example, dangerous objects (eg. hot soup or a knife) are manipulated farther away from the body than fragile objects.

	\item \textbf{Object on hand-selection:} The category of an object can lead to the selection of a particular hand. For example, tools like knives or spoons are typically held with the dominant hand. \cite{Leconte2006,Gonzalez2015} show that the dominant hand crosses the midline more often when a task involves the manipulation of a tool.
	
	\item \textbf{Surface on hand-distance:}
	Surfaces can affect the distance at which a person manipulates objects. For example, handling a pot on a hot stove lead to safety distance to avoid burns. %NO-CITE.  
	
	\item \textbf{Surface on hand-selection:} Spatial layouts can lead to hand selection. For example, a refrigerator's door opens comfortably for the right hand but less for the left. % NO-CITE. 
	
	% [Physical properties]
	\item \textbf{Object-Volume (OV) on hand-selection:}
	The volume of objects can affect hand selection.
	Large objects, but still manageable for a single hand, are likely handled with the more dexterous hand.
	\cite{Whr2018,Seegelke2018} report that the non-dominant hand is for smaller objects while the dominant hand for larger.
	
	\item \textbf{Object-Volume (OV) on hand-distance:} The volume of objects could affect the extent to which a person stretches the arm. For instance, the handle of a frying pan could lead to a near-body grasp. 

	\item \textbf{Surface-Sliding (SS) on hand-distance:} Sliding surfaces can affect the distance at which a person takes an object. For example, when reaching into a sliding drawer, pulling the drawer open could affect hand distance. 
	
	\item \textbf{Surface-Sliding (SS) on hand-selection:} Surfaces that slide can potentially drive hand selection. For example, preferences opening containers with a particular hand and reach into the container with another or the same hand. 

	\item \textbf{Surface-Inside (SC) on hand-distance:} Target objects located inside of containers are harder to reach than those on unconstrained surfaces. For example, reaching for a milk carton stored deep inside the refrigerator demands stretching the arm wide while on a table not. 
	
	\item \textbf{Surface-Inside (SC) on hand-selection:} Targeting objects placed inside of containers can lead  hand-selection. For example, when grabbing an object inside of a drawer, one hand opens the drawer while other reaches into. 
	
	\item \textbf{Hand-dominance (DO) on hand-selection:} The preference of one hand over the other directly influences hand selection. \cite{Bishop1996,Mamolo2004} show that the dominant hand is preferred for close to mid-line targets. 

	\item \textbf{Hand-dominance (DO) on hand-distance:} The preference on a particular hand could lead a person to extend the dominant arm further away then than the non-dominant thereby affecting the proximity at which an object is grasped. 
\end{enumerate}

The mechanization of causal inference \cite{pearl_j:2000a} allows agents to target any of the hypotheses (arrows) embedded in the graph in Figure~\ref{fig:graphical_model}. This article covers one of them to validate the approach. 

The question addressed in this work is whether hand-distance drives hand-selection or not.
This hypothesis involves two decisions people make (unconsciously) when grasping objects with their hands. In other words, the distance at which a person decides to grasp an object (i.e. the arm extension) potentially affects the choice of hand (i.e. left or right).

This question is challenging given that other factors could affect (even simultaneously) the choice of hand. First, hand-selection could be driven by hand-dominance as well as any of the other factors modeled in the graph. Second, inferences have to deal with class imbalances in raw data.
Third, because not all confounding factors are captured by our model, bias-free guarantees are compromised. This is indicated explicitly in the graph by the node of unobserved confounding in  Figure~\ref{fig:graphical_model}. For instance, we have not modeled the mental state of a person, which could play a relevant role in grasping actions and therefore be driving change. 

Note that arrows in causal graphs only state that a potential influence exists, these do not assert certainty. 
Moreover, because causal graphs do not impose any form of functional assumption on the data-generating-process (e.g. linearity), models are expected to transfer across scenarios. For example, the graph in Figure~\ref{fig:graphical_model} is expected to hold true for other actions than the modelled here, e.g. placing. 

In summary, to account for confounding in unconstrained examples of hand manipulation, we propose the model of Figure~\ref{fig:graphical_model}. Among the many hypotheses this graph provides, the causal quantity of hand-distance on hand-selection validates this work. 
Before plugging this model into the machinery of causal inference, the collection of evidence are introduced next.

\subsection{Step 2: Collecting the Evidence}\label{sec:phase2}

A headset and two hand controllers (VR-devices) track user motions in three-dimensional space. 
A game engine (the unreal engine) is responsible for logging streams of data and rendering a near photo-realistic virtual kitchen scenario as shown in Figure 4. The gear users employ to interact with the virtual environment are the HTC Vive Pro and the original handheld controllers.

\begin{figure}[ht]
  \centering
  \includegraphics[width=0.95\linewidth]{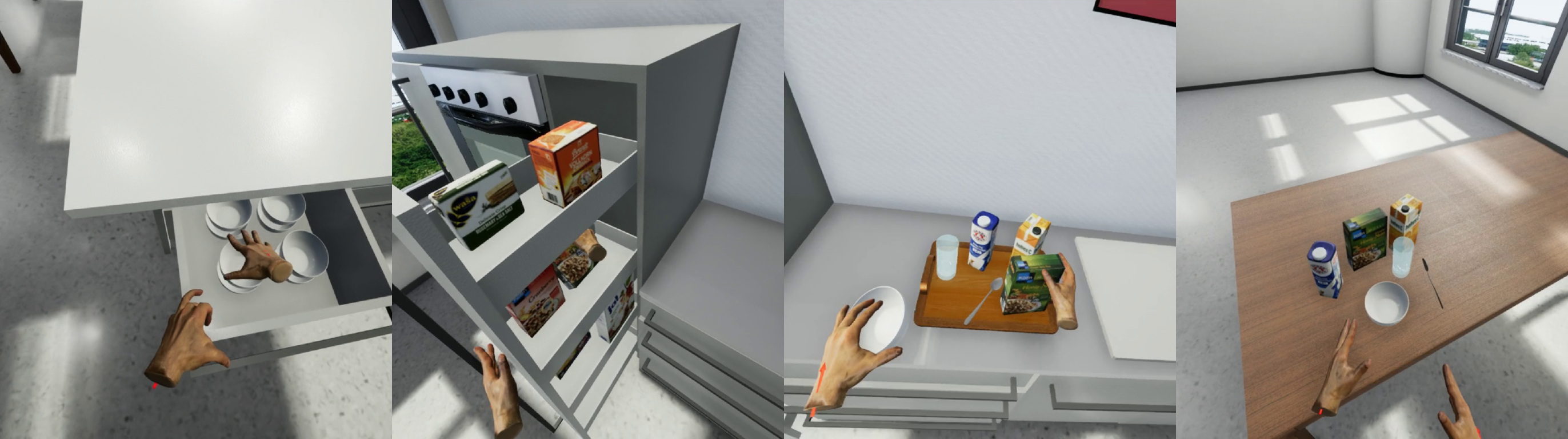}
  \caption[A table setting in the virtual kitchen]{A table setting as experienced by the person in the virtual kitchen. The last image in the sequence shows the table set.} %mention video?
  \label{fig:teaser}
\Description[A table setting in the virtual kitchen]{A table setting as experienced by the person in the virtual kitchen. The last image in the sequence shows the table set.}  
\end{figure}

A headset and two hand controllers (VR devices) track user motions in three-dimensional space. 
A game engine is responsible for logging streams of data and rendering a near photo-realistic virtual kitchen scenario (Figure~\ref{fig:teaser}). The gear users employ to interact with the virtual environment are the HTC Vive Pro and the original handheld controllers. With the tool described in \cite{beetz18knowrob}, raw streams of data are annotated with semantics such as collisions, changes in object states and so on. 
Robots can then access the semantically annotated data with \cite{kazhoyan20vr}.
The focus of this contribution is on the processing of such data. 

A table setting involves a person moving objects from one place to another. This includes searching, transporting, and placing objects in the kitchen - opening and closing containers, and so forth.

The three collections of table settings considered in this work employ the same recording equipment. Likewise, the virtual kitchen layout, the initial position of objects, and the instruction to set the table for breakfast are the same. 
However, each collection of evidence differs considerably. 

Data collection \cite{salinas} was recorded with the purpose to study how people arrange objects on the table (denoted here as Ds-1). The dataset~\cite{haidu19ameva} studies the time spent setting the table under various modalities (in short Ds-2). The collection of data Ds-v validates this work. The key distinction is that Ds-1 and -2 originate from uncontrolled evidence (described in Section~\ref{sec:moreeval}), whereas Ds-v stems from a controlled experiment (discussed in Section~\ref{sec:validation}). 
Note that the unit of analysis are not participants (only 3) but rather grasping actions (order of hundreds).

Hand selection in each set of data is different in spatio-temporal aspects such as the preference of objects selected for the breakfast, the surfaces on which objects are placed, etc. The dashes in the summary Table~\ref{tab:segmenteddataset} are evidence of such asymmetries across data collections. 

\begin{table*} %[ht]
\caption{Summary of three datasets and frequencies of occurrences for different levels of features}\label{tab:segmenteddataset}
\begin{minipage}{\columnwidth}
\begin{center}
\begin{tabular}{lccccc} \toprule
\bfseries User interaction & & & \bfseries Ds-1 & \bfseries Ds-2 & \bfseries Ds-v\\ 
\midrule
\textbf{Object (O):} & \textbf{OV ($m^3$)} & & \textbf{384} 	& \textbf{174} & \textbf{137}\\
Silverware  & $8.2\times10^{-5}$	& & 83 & 70 & 29 \\
Glass    & $1.22\times10^{-3}$  	& & 83 & - 	& 23\\
Milk     & $1.40\times10^{-3}$  	& & 18 & 8 	& 24\\
Juice    & $1.93\times10^{-3}$   & & 88 & - & 21\\
Bowl     & $2.40\times10^{-3}$  	& & 19 & 36 & 16\\
Cereal   & $6.30\times10^{-3}$ 	& & 93 & 28 & 24\\ 
Tray 	 & $9.95\times10^{-3}$ & & - & 32 & - \\ % Tray is an Object AND Surface	
\midrule
\textbf{Surface (S):} 	& \textbf{SC} & \textbf{SS} & \textbf{384} & \textbf{174} & \textbf{137}\\
DiningTable 				& F & F & 12 & 18 & -\\
FrdgArea 				    & T & F & -  & 1 & 1\\
FrdgDrBtmShlf  	    & T & F & 5  & 2 	& 22\\
FrdgGlassShlf 		    & T & T & 41 & - 	& 21\\
IslndArea 				    & F & F & -  & 8 	& 21\\
IslndDrwBtmLft      & T & T & -  & 16 	& -\\
LabFloor   				    & F & F & 5  & - 	& 2\\
OvenArea  				    & F & F & -  & 2 	& -\\
OvenDrwRight 			& T & T & 41 & 16 	& 22\\
SinkArea    			    & F & F & 11 & 24 	& 1\\ % merged with Sink
SnkDrwLftBtm 	    & T & T & -  & - 	& 1\\
SnkDrwLftMid        & T & T & 41 & - 	& 20\\
SnkDrwLftTop		    & T & T  & 41 & 42  & 25\\
Tray 					    & F & F  & 184 & 42 & -\\
%Sink     				& F & F & 3 & - \\ merged to SinkArea
\midrule
\textbf{Hand-selection (H):} & \textbf{DO} & & \textbf{384} & \textbf{174} & \textbf{137}\\
Left  ($H_0$)  & F & & 121 & 96 & 64 \\
Right ($H_1$)  & T & & 263 & 78 & 73 \\
\midrule
\textbf{Hand-distance (D):} & & & \textbf{384} & \textbf{174} & \textbf{137}\\
Close ($\mathit{HD}_0$) & F &  & 192 & 87 & 69 \\
Far   ($\mathit{HD}_1$) & T &  & 192 & 86 & 68 \\
\bottomrule
\end{tabular}
\end{center}
\bigskip
Abbreviation \emph{OV} refers to the volume of an object. \emph{SC} corresponds to surfaces within containers such as the refrigerator's shelf. \emph{SS} refers to surfaces that slide such as drawers. \emph{DO} refers to hand dominance. \emph{SnkDrwLftMid} refers to the middle drawer out of three drawer stacked one over another spatially located left of the sink. \emph{FrdgDrBtmShlf} refers to the bottom shelf on the fridge door. Last, $T$ and $F$ refer to true and false.
\end{minipage}
\end{table*}

Features are either binary, categorical, or numeric as shown in Table~\ref{tab:segmenteddataset}.
The distance of a grasp (denoted by D) is measured when the user triggers a grasp with the motion controllers. Hand-distance is the only feature which requires a transformation because not all selected estimators (introduced in the next section) support continuous treatment values. Discretizing with a median split overcomes this issue.

A median split over the straight-line distance between the tracked headset and handheld VR-devices determines the distance.
The resulting splits for the validation set are $(0.56,0.77)$ meters for a close grasp and $(0.77,0.85)$ meters for a distant grab. Likewise, for Ds-1 the bins corresponding to close/far are $(0.59,0.73)$ and $(0.73,0.92)$. Similarly, the splits for Ds-2 are $(0.47,0.63)$ and $(0.63,0.77)$, correspondingly.
The frequencies for hand-distance are based on these splits as shown in Table~\ref{tab:segmenteddataset}. 

\subsection{Step 3: Drawing Conclusions}\label{sec:phase3} %\label{sec:methods}

To mechanize inferences a Structured Causal Model (SCM) \cite{pearl_j:2000a} is specified. The causal graph defined earlier in step 1 (Figure \ref{fig:graphical_model}) is plugged into the SCM. % M = <U,V,>

Based on the selected hypothesis and on the graph topology, this formalism derives the following (bias-free) expression also known as the estimand \cite{pearl_j:2000a}:
\begin{equation}\label{eq:estimand1}
\begin{split}
\frac{\partial}{\partial D} (Expectation(H|\mathit{O},\mathit{OV},\mathit{S},\mathit{SS},\mathit{SC})) 
\end{split}
\end{equation}

The symbols of expression (\ref{eq:estimand1}) correspond to the features introduced in step 2 (Table \ref{tab:segmenteddataset}). The treatment and outcome in this expression are determined by the hypothesis, i.e. hand-selection ($H$) and hand-distance ($D$). The backdoor criterion \cite{pearl_j:2000a} identifies the remaining symbols as potential factors affecting the hypothesis.

The next element to define is the estimate reported in subsequent sections. The causal estimate is defined by the mean conditional average treatment effect (a.k.a. CATE mean):
\begin{equation}\label{eq:cate}
\begin{split}
E\left[Y_{i, t=1}^{e} - Y_{i, t=0}^{e} | X_i\right] % = \theta_t(X_i)
\end{split}
\end{equation}
where the outcome $Y$ (hand choice) is computed by estimator $e$ (described below) for data point $i$ (instant of time the person triggered a grasp) conditioned on the treatment $t$ (proximity distance) provided the set of observed features $X$ (factors in Figure~\ref{fig:graphical_model}).

Because the data generating process in human hand-behavior is unknown, four estimators operating under differing assumptions cover the full spectrum:~
\begin{itemize}
\item \textbf{linear treatment:} The outcome $Y$ is a linear function of the treatment $X$. Estimator \cite{RePEc:arx:papers:1806.04823} Forest Double Machine Learning Estimator (FDML) imposes this assumption. 
\item \textbf{linear heterogeneity}: The outcome $Y$ is linear on observable characteristics ($Z$). %,$W$). 
Estimator \cite{Bang2005} Linear Doubly Robust Learner (LDRL) 
embeds this assumption.
\item \textbf{linear treatment and heterogeneity:} Estimator \cite{10.1257/aer.p20171038} LinearDML (LDML) 
embeds both of previous assumptions. 
\item \textbf{no assumption:} Estimator \cite{journals/corr/abs-1901-09036} Forest Doubly Robust Learner (FDRL) imposes no operational assumption. 
\end{itemize}
Note that these four are non-parametric estimators suitable for high-dimensional data. Each of them is a composite two-stage ML-algorithm. The final-stage grants desirable properties and is not subject to hyper-parameter optimization unlike the first-stage models, which comprise cross-validated grid-search\footnote{
With parameter grid:
Random Forest Regressor (max\_depth:[3, None], min\_samples\_leaf: [10, 50]),
Gradient Boosting Regressor (n\_estimators: [50, 100], max\_depth:[3], min\_samples\_leaf: [10, 30]),
Random Forest Classifier (max\_depth: [3, 5], min\_samples\_leaf: [10, 50]),
Gradient Boosting Classifier (n\_estimators: [50, 100], max\_depth: [3], min\_samples\_leaf: [10, 30])
.}. 

The selected estimators operate under the unconfoundedness assumption and construct valid confidence intervals. 
Due to small sample size across data sets, statistical significance is set to $0.9$. 
The computation of confidence intervals follows OLS\footnote{Ordinary Least Squares (OLS).} Inference for (LDML and LDRL) and Sub-sampled Honest Forest for (FDML and FDRL).
Further information on these methods are available in \cite{econml}.

\section{Validating the Approach}\label{sec:validation}

A ground truth dataset is recorded to validate the approach. In particular, to verify that the setting derives correct conclusions, a controlled experiment is performed. 

The validation set holds examples of a person setting the table for breakfast. The choice of hand in grasping actions is the only aspect that has been controlled.
The experimenter instructs a person to use either the left, right, or both hands before engaging in the activity. The sequence of instructions is chosen randomly by the experimenter. The case of both hands is a distracting element to avoid suspicion by the participant on the matter being investigated. The data collected with the experiment is summarized under column Ds-v in Table~\ref{tab:segmenteddataset} alongside the other collections of data.

Because hand-selection is randomized in the experiment, only weak (near to null) supporting evidence can be expected for hand-selection on the validation set. 
Therefore, estimators that find support for the hypothesis in the validation data would indicate a red flag on the validity of the setting. On the contrary, ideal results consist of estimators not being able to find evidence in such data. 

After computing the effects on the validation set, four estimators recovered no evidence in support of the hypothesis. This is indicated in Table~\ref{tab:estimates_validation} by the confidence intervals including the zero value for all estimators. The result is clear - the setting recovers the expected effect. 

\begin{table}[htbp] %[ht]
\caption{Effects on the validation set Ds-v}\label{tab:estimates_validation}
\begin{minipage}{\columnwidth}
\begin{center}
\begin{tabular}{lllr}
\toprule
\bfseries Samples & \bfseries Estimator & \bfseries Effect & \bfseries Conf. Intv. \\
\midrule
      137 &   LDML &      -0.04 &   [-0.21, 0.13] \\ %&   0.25 \\
       &      LDRL &      -0.08 &   [-0.25, 0.09] \\ %&   1.40 \\
       &      FDML &       0.01 &   [-0.26, 0.27] \\ %&   0.28 \\
       &      FDRL &      -0.54 &   [-1.57, 0.50] \\ %&  31.89 \\
\bottomrule
\end{tabular}
\end{center}
Four non-parametric estimators report the effect on the validation set comprising 137 grasping actions of a person performing table settings.
The confidence intervals for all estimators include the zero value.
\end{minipage}
\end{table}

This result is encouraging. The setting correctly detected the expected null effect considering the freedom with which the activity had been recorded (free choice in the selection of objects, order, placement, etc., see Table~\ref{tab:segmenteddataset}). 
Moreover, as not all confounding factors are captured by our model, bias cannot be guaranteed to be fully removed. Despite breaking unconfoundedness, the approach manages to recover the correct effect. 
Furthermore, the estimators have been mostly applied to operate at the level of thousands, but in this setting only hundreds of samples are considered. In spite of small quantities of data, the setting is sensible enough to detect the correct effect.

Regarding the methods, Table~\ref{tab:estimates_validation} shows that estimators operating under linear treatment and heterogeneity (top three rows) present narrower confidence intervals than the last variant (bottom row). Smaller confidence intervals for the top three rows makes sense because these estimators impose some structure and thereby deviations are (to some extent) partially restricted. Another reason the least restricted estimator (FDRL) presents the widest intervals is that it requires more data non-forest counterparts. 
However, beyond distinctions, all of them recovered the correct quantity.

Taken together, the approach recovers the expected null effect on the validation set. Moreover, the effect is recoverable with four estimators operating under different assumptions. 

This concludes the validation of the setting on empirical data. 
Next the stability of the model is verified with the validation data, before studying individuals who freely choose their hands (Ds-2 and Ds-3).

\section{Stability of the Approach}\label{sec:robustness}

Models are expected to be applicable beyond scenarios for which they were initially designed. Therefore, it is crucial to verify the stability of the setting under different conditions than the initially planned one. 

Verifying the stability first involves computing baseline values using the original setting. Then, the original setting is modified and the effects are re-computed (e.g. introducing noise to the data, and/or even changing the structure of the model). Ideally, the re-computed effects differ only slightly from the original values. 
In general, the smaller the difference between re-computed and original estimates, the more stable the model is.

Table~\ref{tab:refutation} summarizes the results for three refutation strategies.
The column effect corresponds exactly to the original values presented earlier in Table~\ref{tab:estimates_validation}, while the remaining columns are the re-computed estimates after inflicting changes with a refutation strategy.

\begin{table}%[htbp]
\caption{Refutation strategies using validation data}\label{tab:refutation}
\begin{minipage}{\columnwidth}
\begin{center}
\begin{tabular}{lrrrr}
\toprule
\bfseries Estimator & \bfseries Effect & \bfseries Check-1($\mid$$\Delta$$\mid$) & \bfseries Check-2 & \bfseries Check-3 \\
\midrule
LDML &    -0.04 &    -0.07 (0.03) &    -0.03 (0.01) &  -0.05 (0.01) \\
LDRL &      -0.08 &    -0.11 (0.03) &    -0.04 (0.04) &  -3.43 (3.35) \\
FDML &     0.01 &     0.05 (0.04) &     0.01 (0.0)  &  -0.04 (0.05) \\
FDRL &      -0.54 &    -0.22 (0.32) &    -0.02 (0.52) &  -4.82 (4.28) \\
\bottomrule
\end{tabular}
\end{center}
\bigskip
%\footnotesize\emph{} 
The columns check-1 to -3 correspond to the re-computed effects after changing the setting (i.e. model and/or data). The numbers enclosed by parenthesis ($\mid$$\Delta$$\mid$) are the differences between the original and re-computed effects. Differences close to zero values exhibit robustness against refutations.
\end{minipage}
\end{table}

Refutation strategy check-1 randomizes the treatment and outcome data. Check-2 changes the structure of the model by introducing unobserved random factors that affect the treatment and outcome\footnote{Configured with linear effect strength set to $0.02$ for both treatment and outcome.}. Strategy check-3 computes the estimates on randomly drawn subsets of the original data\footnote{Configured with fraction of data to $0.9$.}. 

The re-computed effects after randomizing the treatment and outcome data (check-1) should not differ from the original values. Indeed this is the case as Table~\ref{tab:refutation} reports near zero differentials except for FDRL. Likewise, when introducing unobserved random confounding factors (check-2) to the causal graph, close to zero $\Delta$ values are reported for the first three estimators except for the last (FDRL). At last, removing random subset of data exhibits near zero differentials for LDML and FDML but not for LDRL and FDRL.

Overall, Table~\ref{tab:refutation} reports that DML estimators outperform DRL across refutation checks as indicated by the near-to-zero differential values ($\mid$$\Delta$$\mid$). DRL, on the other hand, recovers less stable effects. 

DRL typically has a higher variance than DML variants as reported in \cite{econml}, and our results confirm this. Due to imbalances in our datasets, treatment has a small probability of being assigned in the regions of the control space ($Z$) which is critical for DRL estimators, also known as small-overlap. Under such conditions, the DML variants extrapolate better than DRL, and this is the case shown in Table~\ref{tab:refutation}.

\section{Observing Different Individuals}\label{sec:moreeval}

This section analyses the effect of hand-distance on hand-selection for two individuals. Unlike the effects reported in Section~\ref{sec:validation}, in the following subjects exercise the free choice in hand selection - all else remains equal (model, estimators, parameters).

Provided with examples from dataset Ds-1, the estimators recovered positive effects. This is indicated by the confidence intervals not including the zero value for three estimators except for one as shown in Table ~\ref{tab:estimates2}. 
A positive effect means that when targeting objects farther away from the body, the preferred choice is the right hand.

\begin{table}%[htbp] %[ht]
\caption{Effects on observational data Ds-1 and -2}\label{tab:estimates2}
\begin{minipage}{\columnwidth}
\begin{center}
\begin{tabular}{ccccc}
\toprule
\bfseries Data set & \bfseries Smpl. & \bfseries Estimator & \bfseries Effect & \bfseries Conf. Intv. \\ %&  score \\
\midrule
$1$ &      $384$ & LDML &   $0.09$ & $[0.02, 0.17]$  \\ % &   0.25 \\
 &       &      LDRL &       $0.09$ &    $[0.01, 0.16]$  \\ %&   1.40 \\
 &       &    FDML &       $0.16$ &    $[0.05, 0.26]$  \\ %&   0.12 \\
 &       &      FDRL &       $0.08$ &    $[-0.01, 0.17]$ \\ %&   1.04 \\
\hline
$2$ &      $174$ &    LDML &      $-0.13$ &   $[-0.26, 0.01]$ \\ %&   0.26 \\
 &       &      LDRL &      $-0.18$ &   $[-0.42, 0.05]$ \\ %&   3.60 \\
 &       &    FDML &      $-0.13$ &   $[-0.29, 0.02]$ \\ %&   0.26 \\
 &       &      FDRL &      $-0.08$ &   $[-0.31, 0.15]$ \\ %&   2.37 \\
\bottomrule
\end{tabular}
\end{center}
The effect for hand-distance on hand-selection for two individuals freely selecting their hands when setting the table. For one individual the effect is positive, whereas for the other negative. In all cases estimators agree except for FDRL on Ds-1, which is a weak positive effect as the confidence interval includes the null value.
\end{minipage}
\end{table}

% ds-2
In dataset Ds-2 the four estimators derive a weak negative effect. 
This is indicated by the confidence intervals marginally containing the zero. 
Moreover, these intervals are less centered than those reported earlier on the validation set (Table~\ref{tab:estimates_validation}) thus indicating a weaker null effect. A negative effect means that the individual prefers near-to-body grasps with the right hand.

To summarize, different than on the validation set, the resulting effects are not null on evidence which is uncontrolled (Ds-1 and Ds-2). Both individuals differ in the way hand-distance affects hand-selection. One individual engages with the right hand targets that are farther away, whereas the other one prefers nearby.

\section{Explaining Hand Manipulation}\label{sec:interpretation}

So far we have described the effect (direction and strength) that hand-distance has on hand-selection for three individuals. But insights explaining the sensibility of these effects w.r.t factors are missing. 
Interpretability is especially important on inputs for which no ground truth data exists (Ds-1 and Ds-2).

\begin{figure*}
\centering
\subfloat[\label{fig:trees_objvol_1a}\textbf{ds-v:} Objects with bounding boxes smaller than $0.003$ are grasped near the body (red), otherwise at neutral distances (white).]{%
\includegraphics[width=0.49\linewidth]{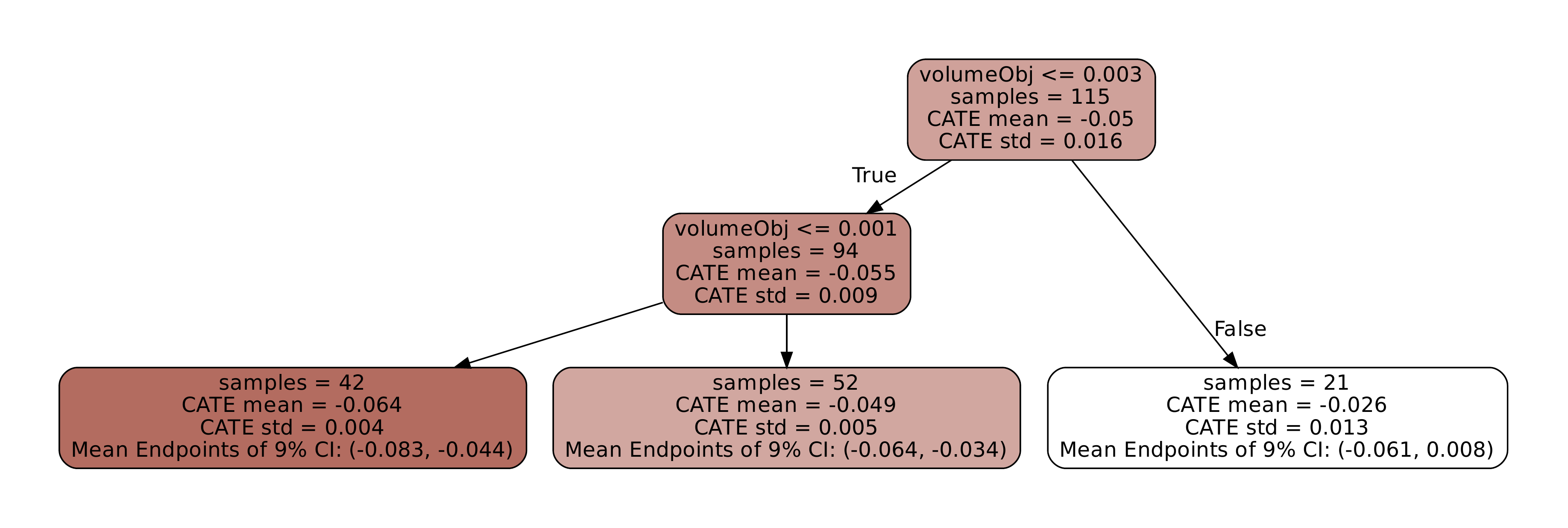}}
\hfill
\subfloat[\label{fig:trees_objvol_1b}\textbf{ds-1:} Objects smaller than volume $0.001$ are grasped near the body, otherwise at a neutral distance.]{%
\includegraphics[width=0.49\linewidth]{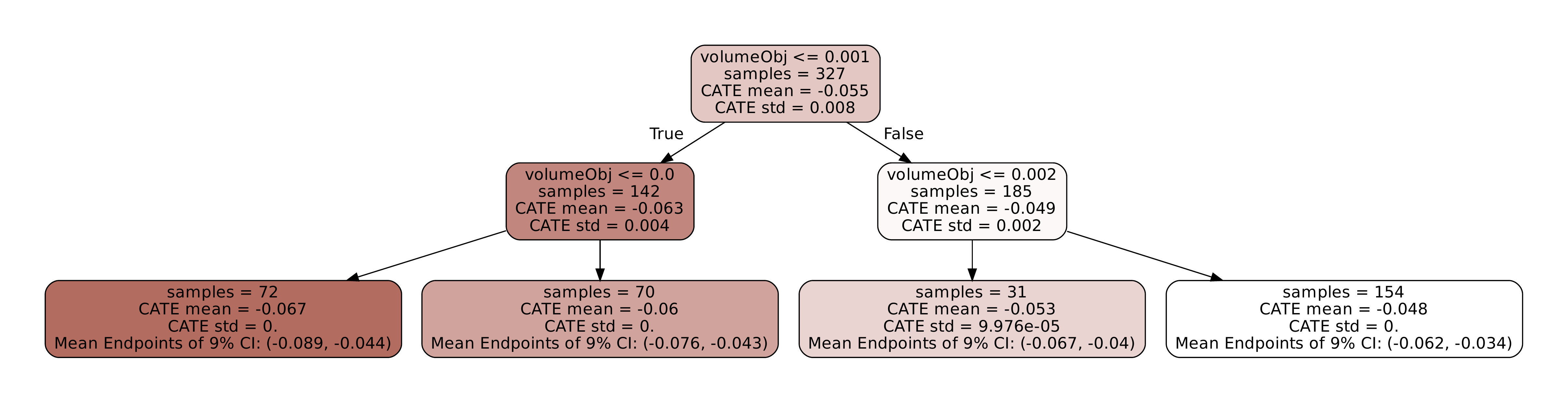}}
\\
\subfloat[\label{fig:trees_objvol_1c}\textbf{ds-2:} Objects with smaller volumes than $0.007$ are targeted closer to the body, otherwise far apart (green).]{%
\includegraphics[width=0.5\linewidth]{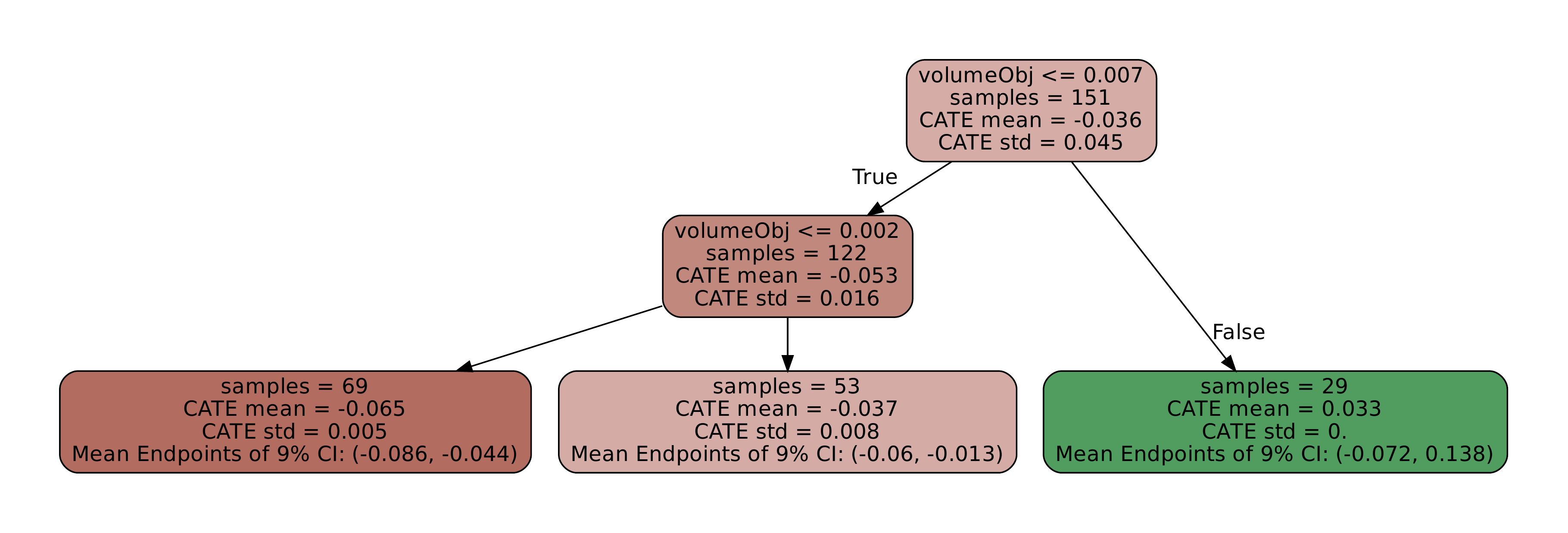}}
\caption[Effect sensitivities]{Interpreting the effect of hand-distance on hand-selection in terms of object volume on each collection of data.}
\label{fig:trees_objvol}
\Description[Effect sensitivities]{Interpreting the effect of hand-distance on hand-selection in terms of object volume on each collection of data.}
\end{figure*}

To this end, the sensibility of the effects (reported earlier in Table~\ref{tab:estimates_validation} and Table~\ref{tab:estimates2}) are interpreted using trees as shown in Figure~\ref{fig:trees_objvol}. 
Every tree interprets the sensibility of a different person. A tree encodes the direction of sensibilities with colors - negative (red), neutral (white) or positive (green) - while the intensity of colors indicates the level of sensibility (either strong or weak). The splitting policy of the tree is in accordance with factors that influence the sensibility of effects. For instance, Figure~\ref{fig:trees_objvol_1a} interprets the effect in terms of object volume (i.e. the bounding box of an object).

Figure~\ref{fig:trees_objvol} shows that the smaller an object is, the closer to the person’s body it is grabbed. This is indicated by the colors ranging from dark red to lighter colors as the volume of objects increases. This behavior is consistent across the three individuals. 
However, the quantitative value of when an object is considered to be big/small differs for each individual as indicated by the root node of the trees.

% summary
To summarize, the effects presented in earlier sections (Table~\ref{tab:estimates_validation} and Table~\ref{tab:estimates2}) are explainable in terms of sensibility to influential factors.
The sensibility of the effect that hand-proximity exerts on hand-selection are explained in terms of object-volume. 
Different individuals present consistent trends but differ in terms of quantitative preferences of object volumes. 
Individualized insights like these have the potential to enable robots to provide personalized assistance. 

\section{Related Work}

Autonomous robots could one day spare the cost of programming them. One way to reduce the amount of programming is by providing robots with examples of the task space \cite{journals/ras/ArgallCVB09}. 
Another strategy is learning the control space showing exemplary movements to the robots  \cite{journals/ijrr/KulicOLIN12}. 
Providing robots with examples instead of code shifts the burden from programming to teaching them. Teaching is demanding as high quality, successful, and sufficient examples are required.

One strategy to cut the cost of teaching robots, is to learn by doing in simulation environments as done in \cite{journals/corr/GoodfellowPMXWOCB14,Silver17a}. Virtual settings offer nearly endless synthetic examples for an initial upfront cost. Computational resources leverage exhaustive explorations in simulation space. Besides sample inefficiency, solutions from simulated environments must overcome transfer issues to real-world scenarios \cite{hofer2020perspectives}.

Unlike the previous strategies, children approach things differently. They can learn quickly from a few `uncurated' good-and-bad examples \cite{Guerin11}. Moreover, children can explain their answers \cite{Walker2016}, an ability often overlooked in model-free data-driven approaches. More crucially, children transfer knowledge from one domain to another \cite{Olh2016}.

Developmental AI and Robotics are still in their infancy \cite{Eppe2021}. A key achievement in recent years has been the algorithmization of causal inference \cite{pearl_j:2000a}. Such advancements offer opportunities to explore alternative reasoning capabilities for robots. 

This contribution explores the feasibility to equip robot with child-like observational abilities. To prove the concept, we setup a framework and confront the observational capabilities to a few uncurated empirical examples of hand manipulation.

The authors in \cite{journals/tist/FireZ16} build a causal and-or graph for objects and actions from videos. Unlike our contribution, this work focus on action detection and automating the discovery of causal models at the expense of transferability and interpretability.

\cite{journals/corr/abs-1903-01267} employs SCM to recover trajectories for different kinesthetic teaching styles. However, this work differs considerably. First, the end-effector of a robot is modeled whereas our contribution involves the human hand. Second, the authors address robot poses while our contribution targets human motor decisions. Third, their source of evidence is structured and curated when compared to the unconstrained activities considered here.

\cite{uhde_robot_iros2020} employed causal inference to reduce the space of action pairs recorded in a virtual environment thereby boosting performance by six fold.
This work puts the focus on causal dependencies of action sequences rather than on decision-making and context as presented here.

%%%%%%%%%%%%%%%%%%%%%%%%%%%%%%%%%%%%%%%%%%%%%%%%%%%%%%%%%%%%%%%%%%%%%%%%

\section{Conclusion}

To unlock individualized robotic support in human-robot interactions,
this contribution explores the validity of causal quantities on empirical evidence in the context of everyday hand manipulations. 

The conditions under which the approach is examined are challenging. First, this contribution does not include all relevant factors in the model thus breaking the unconfoundedness assumption. Therefore, we can expect bias-reduced estimates but not bias-free. This setting is valuable to investigate because children cope with states of incompleteness. 
Second, the non-parametric estimators of this contribution typically target samples in the order of thousands but only hundreds are considered here.

The approach recovered the null effect on the validation data. Four estimators operating under differing assumptions agreed on this result. Moreover, the model is stable against three refutation strategies on two collections of data. 

Unlike the validation set, targeting observational data does not derive a null effect. Instead, estimates result in positive and weak negative estimates for two individuals.

Taken both results, the approach is capable to detect null and non-null effects. The possibility of the framework being either too sensitive (to derive null effects) or insensitive (to detect non-null effects) is thus excluded. Moreover, the opposing estimates derived for two individuals manifest the feasibility to recover effects in different directions. 

Lastly, we interpret the sensibility of effects w.r.t. contextual factors (i.e. object-volume) potentially affecting hand behavior (i.e. the relationship between hand-distance and hand-selection). The trees highlighted commonalities as well as individualities across three subjects. Such as the preference of the right hand for distant targets, and the personal threshold of objects considered small or large given by the root of the tree.

Our results are encouraging and foundational for further research.
In the future, we plan to study other actions such as placing, opening, and closing. We look forward to extending our model with factors such as shape, handles, grasping points, and full-body tracking.

%%%%%%%%%%%%%%%%%%%%%%%%%%%%%%%%%%%%%%%%%%%%%%%%%%%%%%%%%%%%%%%%%%%%%%%%

%%% The acknowledgments section is defined using the "acks" environment
%%% (rather than an unnumbered section). The use of this environment 
%%% ensures the proper identification of the section in the article 
%%% metadata as well as the consistent spelling of the heading.

\begin{acks}
% If you wish to include any acknowledgments in your paper (e.g., to 
% people or funding agencies), please do so using the `\texttt{acks}' 
% environment. Note that the text of your acknowledgments will be omitted
% if you compile your document with the `\texttt{anonymous}' option.
This work is partially funded by the Scholarship Programme Don Carlos Antonio L\'opez (BECAL) and DFG, as part of Collaborative Research Center (Sonderforschungsbereich) 1320 EASE - Everyday Activity Science and Engineering, University of Bremen (http://www.ease-crc.org/).
\end{acks}

%%%%%%%%%%%%%%%%%%%%%%%%%%%%%%%%%%%%%%%%%%%%%%%%%%%%%%%%%%%%%%%%%%%%%%%%

%%% The next two lines define, first, the bibliography style to be 
%%% applied, and, second, the bibliography file to be used.

\bibliographystyle{ACM-Reference-Format} 
\bibliography{sample}


\begin{thebibliography}{37}

%%% the Web address.

\ifx \showCODEN    \undefined \def \showCODEN     #1{\unskip}     \fi
\ifx \showDOI      \undefined \def \showDOI       #1{#1}\fi
\ifx \showISBNx    \undefined \def \showISBNx     #1{\unskip}     \fi
\ifx \showISBNxiii \undefined \def \showISBNxiii  #1{\unskip}     \fi
\ifx \showISSN     \undefined \def \showISSN      #1{\unskip}     \fi
\ifx \showLCCN     \undefined \def \showLCCN      #1{\unskip}     \fi
\ifx \shownote     \undefined \def \shownote      #1{#1}          \fi
\ifx \showarticletitle \undefined \def \showarticletitle #1{#1}   \fi
\ifx \showURL      \undefined \def \showURL       {\relax}        \fi
% invisible to TeX
\providecommand\bibfield[2]{#2}
\providecommand\bibinfo[2]{#2}
\providecommand\natexlab[1]{#1}
\providecommand\showeprint[2][]{arXiv:#2}

\bibitem[\protect\citeauthoryear{Angelov, Hristov, and Ramamoorthy}{Angelov
  et~al\mbox{.}}{2019}]%
        {journals/corr/abs-1903-01267}
\bibfield{author}{\bibinfo{person}{Daniel Angelov}, \bibinfo{person}{Yordan
  Hristov}, {and} \bibinfo{person}{Subramanian Ramamoorthy}.}
  \bibinfo{year}{2019}\natexlab{}.
\newblock \showarticletitle{Using Causal Analysis to Learn Specifications from
  Task Demonstrations}.
\newblock \bibinfo{journal}{\emph{CoRR}}  \bibinfo{volume}{abs/1903.01267}
  (\bibinfo{year}{2019}).
\newblock
\urldef\tempurl%
\url{http://arxiv.org/abs/1903.01267}
\showURL{%
\tempurl}


\bibitem[\protect\citeauthoryear{Argall, Chernova, Veloso, and Browning}{Argall
  et~al\mbox{.}}{2009}]%
        {journals/ras/ArgallCVB09}
\bibfield{author}{\bibinfo{person}{Brenna~D. Argall}, \bibinfo{person}{Sonia
  Chernova}, \bibinfo{person}{Manuela~M. Veloso}, {and} \bibinfo{person}{Brett
  Browning}.} \bibinfo{year}{2009}\natexlab{}.
\newblock \showarticletitle{A survey of robot learning from demonstration}.
\newblock \bibinfo{journal}{\emph{Robotics and Autonomous Systems}}
  \bibinfo{volume}{57}, \bibinfo{number}{5} (\bibinfo{year}{2009}),
  \bibinfo{pages}{469--483}.
\newblock


\bibitem[\protect\citeauthoryear{Bang and Robins}{Bang and Robins}{2005}]%
        {Bang2005}
\bibfield{author}{\bibinfo{person}{Heejung Bang} {and}
  \bibinfo{person}{James~M. Robins}.} \bibinfo{year}{2005}\natexlab{}.
\newblock \showarticletitle{Doubly Robust Estimation in Missing Data and Causal
  Inference Models}.
\newblock \bibinfo{journal}{\emph{Biometrics}} \bibinfo{volume}{61},
  \bibinfo{number}{4} (\bibinfo{date}{June} \bibinfo{year}{2005}),
  \bibinfo{pages}{962--973}.
\newblock
\urldef\tempurl%
\url{https://doi.org/10.1111/j.1541-0420.2005.00377.x}
\showDOI{\tempurl}


\bibitem[\protect\citeauthoryear{Beetz, Be{\ss}ler, Haidu, Pomarlan, Bozcuoglu,
  and Bartels}{Beetz et~al\mbox{.}}{2018}]%
        {beetz18knowrob}
\bibfield{author}{\bibinfo{person}{Michael Beetz}, \bibinfo{person}{Daniel
  Be{\ss}ler}, \bibinfo{person}{Andrei Haidu}, \bibinfo{person}{Mihai
  Pomarlan}, \bibinfo{person}{Asil~Kaan Bozcuoglu}, {and}
  \bibinfo{person}{Georg Bartels}.} \bibinfo{year}{2018}\natexlab{}.
\newblock \showarticletitle{KnowRob 2.0 -- A 2nd Generation Knowledge
  Processing Framework for Cognition-enabled Robotic Agents}. In
  \bibinfo{booktitle}{\emph{International Conference on Robotics and Automation
  (ICRA)}}. \bibinfo{address}{Brisbane, Australia}.
\newblock


\bibitem[\protect\citeauthoryear{Bishop, Ross, Daniels, and Bright}{Bishop
  et~al\mbox{.}}{1996}]%
        {Bishop1996}
\bibfield{author}{\bibinfo{person}{D.~V.~M. Bishop}, \bibinfo{person}{V.~A.
  Ross}, \bibinfo{person}{M.~S. Daniels}, {and} \bibinfo{person}{P. Bright}.}
  \bibinfo{year}{1996}\natexlab{}.
\newblock \showarticletitle{The measurement of hand preference: A validation
  study comparing three groups of right-handers}.
\newblock \bibinfo{journal}{\emph{British Journal of Psychology}}
  \bibinfo{volume}{87}, \bibinfo{number}{2} (\bibinfo{date}{\#may\#}
  \bibinfo{year}{1996}), \bibinfo{pages}{269--285}.
\newblock
\urldef\tempurl%
\url{https://doi.org/10.1111/j.2044-8295.1996.tb02590.x}
\showDOI{\tempurl}


\bibitem[\protect\citeauthoryear{Chernozhukov, Chetverikov, Demirer, Duflo,
  Hansen, and Newey}{Chernozhukov et~al\mbox{.}}{2017}]%
        {10.1257/aer.p20171038}
\bibfield{author}{\bibinfo{person}{Victor Chernozhukov}, \bibinfo{person}{Denis
  Chetverikov}, \bibinfo{person}{Mert Demirer}, \bibinfo{person}{Esther Duflo},
  \bibinfo{person}{Christian Hansen}, {and} \bibinfo{person}{Whitney Newey}.}
  \bibinfo{year}{2017}\natexlab{}.
\newblock \showarticletitle{Double/Debiased/Neyman Machine Learning of
  Treatment Effects}.
\newblock \bibinfo{journal}{\emph{American Economic Review}}
  \bibinfo{volume}{107}, \bibinfo{number}{5} (\bibinfo{date}{May}
  \bibinfo{year}{2017}), \bibinfo{pages}{261--65}.
\newblock
\urldef\tempurl%
\url{https://doi.org/10.1257/aer.p20171038}
\showDOI{\tempurl}


\bibitem[\protect\citeauthoryear{Chernozhukov, Nekipelov, Semenova, and
  Syrgkanis}{Chernozhukov et~al\mbox{.}}{2018}]%
        {RePEc:arx:papers:1806.04823}
\bibfield{author}{\bibinfo{person}{Victor Chernozhukov}, \bibinfo{person}{Denis
  Nekipelov}, \bibinfo{person}{Vira Semenova}, {and} \bibinfo{person}{Vasilis
  Syrgkanis}.} \bibinfo{year}{2018}\natexlab{}.
\newblock \bibinfo{booktitle}{\emph{{Plug-in Regularized Estimation of
  High-Dimensional Parameters in Nonlinear Semiparametric Models}}}.
\newblock \bibinfo{type}{Papers} 1806.04823. \bibinfo{institution}{arXiv.org}.
\newblock
\urldef\tempurl%
\url{https://ideas.repec.org/p/arx/papers/1806.04823.html}
\showURL{%
\tempurl}


\bibitem[\protect\citeauthoryear{Eppe, Wermter, Hafner, and Nagai}{Eppe
  et~al\mbox{.}}{2021}]%
        {Eppe2021}
\bibfield{author}{\bibinfo{person}{Manfred Eppe}, \bibinfo{person}{Stefan
  Wermter}, \bibinfo{person}{Verena~V. Hafner}, {and} \bibinfo{person}{Yukie
  Nagai}.} \bibinfo{year}{2021}\natexlab{}.
\newblock \showarticletitle{Developmental Robotics and its Role Towards
  Artificial General Intelligence}.
\newblock \bibinfo{journal}{\emph{{KI} - K\"{u}nstliche Intelligenz}}
  \bibinfo{volume}{35}, \bibinfo{number}{1} (\bibinfo{date}{Feb.}
  \bibinfo{year}{2021}), \bibinfo{pages}{5--7}.
\newblock
\urldef\tempurl%
\url{https://doi.org/10.1007/s13218-021-00706-w}
\showDOI{\tempurl}


\bibitem[\protect\citeauthoryear{Fire and Zhu}{Fire and Zhu}{2016}]%
        {journals/tist/FireZ16}
\bibfield{author}{\bibinfo{person}{Amy~Sue Fire} {and}
  \bibinfo{person}{Song-Chun Zhu}.} \bibinfo{year}{2016}\natexlab{}.
\newblock \showarticletitle{Learning Perceptual Causality from Video}.
\newblock \bibinfo{journal}{\emph{ACM TIST}} \bibinfo{volume}{7},
  \bibinfo{number}{2} (\bibinfo{year}{2016}), \bibinfo{pages}{23:1--23:22}.
\newblock


\bibitem[\protect\citeauthoryear{Foster and Syrgkanis}{Foster and
  Syrgkanis}{2019}]%
        {journals/corr/abs-1901-09036}
\bibfield{author}{\bibinfo{person}{Dylan~J. Foster} {and}
  \bibinfo{person}{Vasilis Syrgkanis}.} \bibinfo{year}{2019}\natexlab{}.
\newblock \showarticletitle{Orthogonal Statistical Learning}.
\newblock \bibinfo{journal}{\emph{CoRR}}  \bibinfo{volume}{abs/1901.09036}
  (\bibinfo{year}{2019}).
\newblock
\urldef\tempurl%
\url{http://arxiv.org/abs/1901.09036}
\showURL{%
\tempurl}


\bibitem[\protect\citeauthoryear{Gabbard and Helbig}{Gabbard and
  Helbig}{2004}]%
        {Gabbard2004}
\bibfield{author}{\bibinfo{person}{Carl Gabbard} {and}
  \bibinfo{person}{Casi~Rabb Helbig}.} \bibinfo{year}{2004}\natexlab{}.
\newblock \showarticletitle{What drives children?s limb selection for reaching
  in hemispace?}
\newblock \bibinfo{journal}{\emph{Experimental Brain Research}}
  \bibinfo{volume}{156}, \bibinfo{number}{3} (\bibinfo{date}{June}
  \bibinfo{year}{2004}), \bibinfo{pages}{325--332}.
\newblock
\urldef\tempurl%
\url{https://doi.org/10.1007/s00221-003-1792-y}
\showDOI{\tempurl}


\bibitem[\protect\citeauthoryear{Gabbard and Rabb}{Gabbard and Rabb}{2000}]%
        {Gabbard2000}
\bibfield{author}{\bibinfo{person}{Carl Gabbard} {and} \bibinfo{person}{Casi
  Rabb}.} \bibinfo{year}{2000}\natexlab{}.
\newblock \showarticletitle{What Determines Choice of Limb for Unimanual
  Reaching Movements?}
\newblock \bibinfo{journal}{\emph{The Journal of General Psychology}}
  \bibinfo{volume}{127}, \bibinfo{number}{2} (\bibinfo{date}{\#apr\#}
  \bibinfo{year}{2000}), \bibinfo{pages}{178--184}.
\newblock
\urldef\tempurl%
\url{https://doi.org/10.1080/00221300009598577}
\showDOI{\tempurl}


\bibitem[\protect\citeauthoryear{Gonzalez, Flindall, and Stone}{Gonzalez
  et~al\mbox{.}}{2015}]%
        {Gonzalez2015}
\bibfield{author}{\bibinfo{person}{Claudia L.~R. Gonzalez},
  \bibinfo{person}{Jason~W. Flindall}, {and} \bibinfo{person}{Kayla~D. Stone}.}
  \bibinfo{year}{2015}\natexlab{}.
\newblock \showarticletitle{Hand preference across the lifespan: effects of
  end-goal, task nature, and object location}.
\newblock \bibinfo{journal}{\emph{Frontiers in Psychology}}
  \bibinfo{volume}{5} (\bibinfo{date}{\#jan\#} \bibinfo{year}{2015}).
\newblock
\urldef\tempurl%
\url{https://doi.org/10.3389/fpsyg.2014.01579}
\showDOI{\tempurl}


\bibitem[\protect\citeauthoryear{Goodfellow, Pouget-Abadie, Mirza, Xu,
  Warde-Farley, Ozair, Courville, and Bengio}{Goodfellow et~al\mbox{.}}{2014}]%
        {journals/corr/GoodfellowPMXWOCB14}
\bibfield{author}{\bibinfo{person}{Ian Goodfellow}, \bibinfo{person}{Jean
  Pouget-Abadie}, \bibinfo{person}{Mehdi Mirza}, \bibinfo{person}{Bing Xu},
  \bibinfo{person}{David Warde-Farley}, \bibinfo{person}{Sherjil Ozair},
  \bibinfo{person}{Aaron Courville}, {and} \bibinfo{person}{Yoshua Bengio}.}
  \bibinfo{year}{2014}\natexlab{}.
\newblock \showarticletitle{Generative Adversarial Nets}. In
  \bibinfo{booktitle}{\emph{Advances in Neural Information Processing
  Systems}}, \bibfield{editor}{\bibinfo{person}{Z.~Ghahramani},
  \bibinfo{person}{M.~Welling}, \bibinfo{person}{C.~Cortes},
  \bibinfo{person}{N.~Lawrence}, {and} \bibinfo{person}{K.~Q. Weinberger}}
  (Eds.), Vol.~\bibinfo{volume}{27}. \bibinfo{publisher}{Curran Associates,
  Inc.}
\newblock
\urldef\tempurl%
\url{https://proceedings.neurips.cc/paper/2014/file/5ca3e9b122f61f8f06494c97b1afccf3-Paper.pdf}
\showURL{%
\tempurl}


\bibitem[\protect\citeauthoryear{Gopnik}{Gopnik}{2010}]%
        {Gopnik2010}
\bibfield{author}{\bibinfo{person}{Alison Gopnik}.}
  \bibinfo{year}{2010}\natexlab{}.
\newblock \showarticletitle{How Babies Think}.
\newblock \bibinfo{journal}{\emph{Scientific American}} \bibinfo{volume}{303},
  \bibinfo{number}{1} (\bibinfo{date}{July} \bibinfo{year}{2010}),
  \bibinfo{pages}{76--81}.
\newblock
\urldef\tempurl%
\url{https://doi.org/10.1038/scientificamerican0710-76}
\showDOI{\tempurl}


\bibitem[\protect\citeauthoryear{Guerin}{Guerin}{2011}]%
        {Guerin11}
\bibfield{author}{\bibinfo{person}{Frank Guerin}.}
  \bibinfo{year}{2011}\natexlab{}.
\newblock \showarticletitle{Learning like a baby: a survey of artificial
  intelligence approaches}.
\newblock \bibinfo{journal}{\emph{Knowledge Eng. Review}} \bibinfo{volume}{26},
  \bibinfo{number}{2} (\bibinfo{year}{2011}), \bibinfo{pages}{209--236}.
\newblock


\bibitem[\protect\citeauthoryear{Haidu and Beetz}{Haidu and Beetz}{2019}]%
        {haidu19ameva}
\bibfield{author}{\bibinfo{person}{Andrei Haidu} {and} \bibinfo{person}{Michael
  Beetz}.} \bibinfo{year}{2019}\natexlab{}.
\newblock \showarticletitle{Automated Models of Human Everyday Activity based
  on Game and Virtual Reality Technology}. In
  \bibinfo{booktitle}{\emph{International Conference on Robotics and Automation
  (ICRA)}}. \bibinfo{address}{Montreal, Canada}.
\newblock


\bibitem[\protect\citeauthoryear{Höfer, Bekris, Handa, Gamboa, Golemo,
  Mozifian, Atkeson, Fox, Goldberg, Leonard, Liu, Peters, Song, Welinder, and
  White}{Höfer et~al\mbox{.}}{2020}]%
        {hofer2020perspectives}
\bibfield{author}{\bibinfo{person}{Sebastian Höfer}, \bibinfo{person}{Kostas
  Bekris}, \bibinfo{person}{Ankur Handa}, \bibinfo{person}{Juan~Camilo Gamboa},
  \bibinfo{person}{Florian Golemo}, \bibinfo{person}{Melissa Mozifian},
  \bibinfo{person}{Chris Atkeson}, \bibinfo{person}{Dieter Fox},
  \bibinfo{person}{Ken Goldberg}, \bibinfo{person}{John Leonard},
  \bibinfo{person}{C.~Karen Liu}, \bibinfo{person}{Jan Peters},
  \bibinfo{person}{Shuran Song}, \bibinfo{person}{Peter Welinder}, {and}
  \bibinfo{person}{Martha White}.} \bibinfo{year}{2020}\natexlab{}.
\newblock \bibinfo{title}{Perspectives on Sim2Real Transfer for Robotics: A
  Summary of the R:SS 2020 Workshop}.
\newblock
\newblock
\showeprint[arxiv]{2012.03806}~[cs.RO]


\bibitem[\protect\citeauthoryear{Kazhoyan, Hawkin, Koralewski, Haidu, and
  Beetz}{Kazhoyan et~al\mbox{.}}{2020}]%
        {kazhoyan20vr}
\bibfield{author}{\bibinfo{person}{Gayane Kazhoyan}, \bibinfo{person}{Alina
  Hawkin}, \bibinfo{person}{Sebastian Koralewski}, \bibinfo{person}{Andrei
  Haidu}, {and} \bibinfo{person}{Michael Beetz}.}
  \bibinfo{year}{2020}\natexlab{}.
\newblock \showarticletitle{Learning Motion Parameterizations of Mobile Pick
  and Place Actions from Observing Humans in Virtual Environments}. In
  \bibinfo{booktitle}{\emph{IEEE/RSJ International Conference on Intelligent
  Robots and Systems (IROS)}}.
\newblock
\urldef\tempurl%
\url{https://doi.org/10.1109/IROS45743.2020.9341458}
\showDOI{\tempurl}


\bibitem[\protect\citeauthoryear{Kock}{Kock}{2015}]%
        {journals/ijec/Kock15}
\bibfield{author}{\bibinfo{person}{Ned Kock}.} \bibinfo{year}{2015}\natexlab{}.
\newblock \showarticletitle{How Likely is Simpson's Paradox in Path Models?}
\newblock \bibinfo{journal}{\emph{IJeC}} \bibinfo{volume}{11},
  \bibinfo{number}{1} (\bibinfo{year}{2015}), \bibinfo{pages}{1--7}.
\newblock


\bibitem[\protect\citeauthoryear{Kulic, Ott, Lee, Ishikawa, and Nakamura}{Kulic
  et~al\mbox{.}}{2012}]%
        {journals/ijrr/KulicOLIN12}
\bibfield{author}{\bibinfo{person}{Dana Kulic}, \bibinfo{person}{Christian
  Ott}, \bibinfo{person}{Dongheui Lee}, \bibinfo{person}{Junichi Ishikawa},
  {and} \bibinfo{person}{Yoshihiko Nakamura}.} \bibinfo{year}{2012}\natexlab{}.
\newblock \showarticletitle{Incremental learning of full body motion primitives
  and their sequencing through human motion observation}.
\newblock \bibinfo{journal}{\emph{I. J. Robotics Res}} \bibinfo{volume}{31},
  \bibinfo{number}{3} (\bibinfo{year}{2012}), \bibinfo{pages}{330--345}.
\newblock


\bibitem[\protect\citeauthoryear{Leconte and Fagard}{Leconte and
  Fagard}{2006}]%
        {Leconte2006}
\bibfield{author}{\bibinfo{person}{Pascale Leconte} {and}
  \bibinfo{person}{Jacqueline Fagard}.} \bibinfo{year}{2006}\natexlab{}.
\newblock \showarticletitle{Which factors affect hand selection in children's
  grasping in hemispace? Combined effects of task demand and motor dominance}.
\newblock \bibinfo{journal}{\emph{Brain and Cognition}} \bibinfo{volume}{60},
  \bibinfo{number}{1} (\bibinfo{date}{\#feb\#} \bibinfo{year}{2006}),
  \bibinfo{pages}{88--93}.
\newblock
\urldef\tempurl%
\url{https://doi.org/10.1016/j.bandc.2005.09.009}
\showDOI{\tempurl}


\bibitem[\protect\citeauthoryear{Mamolo, Roy, Bryden, and Rohr}{Mamolo
  et~al\mbox{.}}{2004}]%
        {Mamolo2004}
\bibfield{author}{\bibinfo{person}{Carla~M. Mamolo}, \bibinfo{person}{Eric~A.
  Roy}, \bibinfo{person}{Pamela~J. Bryden}, {and} \bibinfo{person}{Linda~E.
  Rohr}.} \bibinfo{year}{2004}\natexlab{}.
\newblock \showarticletitle{The effects of skill demands and object position on
  the distribution of preferred hand reaches}.
\newblock \bibinfo{journal}{\emph{Brain and Cognition}} \bibinfo{volume}{55},
  \bibinfo{number}{2} (\bibinfo{date}{\#jul\#} \bibinfo{year}{2004}),
  \bibinfo{pages}{349--351}.
\newblock
\urldef\tempurl%
\url{https://doi.org/10.1016/j.bandc.2004.02.041}
\showDOI{\tempurl}


\bibitem[\protect\citeauthoryear{Ol{\'{a}}h, Elekes, Pet{\H{o}}, Peres, and
  Kir{\'{a}}ly}{Ol{\'{a}}h et~al\mbox{.}}{2016}]%
        {Olh2016}
\bibfield{author}{\bibinfo{person}{Katalin Ol{\'{a}}h},
  \bibinfo{person}{Fruzsina Elekes}, \bibinfo{person}{R{\'{e}}ka Pet{\H{o}}},
  \bibinfo{person}{Krisztina Peres}, {and} \bibinfo{person}{Ildik{\'{o}}
  Kir{\'{a}}ly}.} \bibinfo{year}{2016}\natexlab{}.
\newblock \showarticletitle{3-Year-Old Children Selectively Generalize Object
  Functions Following a Demonstration from a Linguistic In-group Member:
  Evidence from the Phenomenon of Scale Error}.
\newblock \bibinfo{journal}{\emph{Frontiers in Psychology}}
  \bibinfo{volume}{7} (\bibinfo{date}{June} \bibinfo{year}{2016}).
\newblock
\urldef\tempurl%
\url{https://doi.org/10.3389/fpsyg.2016.00963}
\showDOI{\tempurl}


\bibitem[\protect\citeauthoryear{Pavlides and Perlman}{Pavlides and
  Perlman}{2009}]%
        {Pavlides:2009:HLS}
\bibfield{author}{\bibinfo{person}{Marios~G. Pavlides} {and}
  \bibinfo{person}{Michael~D. Perlman}.} \bibinfo{year}{2009}\natexlab{}.
\newblock \showarticletitle{How Likely Is {Simpson's Paradox}?}
\newblock \bibinfo{journal}{\emph{The American Statistician}}
  \bibinfo{volume}{63}, \bibinfo{number}{3} (\bibinfo{date}{\#aug\#}
  \bibinfo{year}{2009}), \bibinfo{pages}{226--233}.
\newblock
\showCODEN{ASTAAJ}
\showISSN{0003-1305 (print), 1537-2731 (electronic)}
\urldef\tempurl%
\url{https://doi.org/10.1198/tast.2009.09007}
\showDOI{\tempurl}


\bibitem[\protect\citeauthoryear{Pearl}{Pearl}{2000}]%
        {pearl_j:2000a}
\bibfield{author}{\bibinfo{person}{Judea Pearl}.}
  \bibinfo{year}{2000}\natexlab{}.
\newblock \bibinfo{booktitle}{\emph{Causality: Models, Reasoning, and
  Inference}}.
\newblock \bibinfo{publisher}{Cambridge University Press},
  \bibinfo{address}{Cambridge, England}.
\newblock
\showISBNx{0-521-77362-8}


\bibitem[\protect\citeauthoryear{Pearl}{Pearl}{2014}]%
        {Pearl2014}
\bibfield{author}{\bibinfo{person}{Judea Pearl}.}
  \bibinfo{year}{2014}\natexlab{}.
\newblock \showarticletitle{Comment: Understanding Simpson's Paradox}.
\newblock \bibinfo{journal}{\emph{The American Statistician}}
  \bibinfo{volume}{68}, \bibinfo{number}{1} (\bibinfo{date}{Jan.}
  \bibinfo{year}{2014}), \bibinfo{pages}{8--13}.
\newblock
\urldef\tempurl%
\url{https://doi.org/10.1080/00031305.2014.876829}
\showDOI{\tempurl}


\bibitem[\protect\citeauthoryear{Pinacho, Wich, Yazdani, and Beetz}{Pinacho
  et~al\mbox{.}}{2018}]%
        {salinas}
\bibfield{author}{\bibinfo{person}{Lisset~Salinas Pinacho},
  \bibinfo{person}{Alexander Wich}, \bibinfo{person}{Fereshta Yazdani}, {and}
  \bibinfo{person}{Michael Beetz}.} \bibinfo{year}{2018}\natexlab{}.
\newblock \showarticletitle{Acquiring Knowledge of Object Arrangements from
  Human Examples for Household Robots}. In \bibinfo{booktitle}{\emph{KI 2018:
  Advances in Artificial Intelligence: 41st German Conference on AI, Berlin,
  Germany, September 24--28, 2018, Proceedings}}, Vol.~\bibinfo{volume}{11117}.
  Springer, \bibinfo{pages}{131}.
\newblock
\urldef\tempurl%
\url{https://link.springer.com/content/pdf/10.1007%2F978-3-030-00111-7_12.pdf}
\showURL{%
\tempurl}


\bibitem[\protect\citeauthoryear{Research}{Research}{2019}]%
        {econml}
\bibfield{author}{\bibinfo{person}{Microsoft Research}.}
  \bibinfo{year}{2019}\natexlab{}.
\newblock \bibinfo{title}{{EconML}: {A Python Package for ML-Based
  Heterogeneous Treatment Effects Estimation}}.
\newblock \bibinfo{howpublished}{https://github.com/microsoft/EconML}.
\newblock
\newblock
\shownote{Version 0.x.}


\bibitem[\protect\citeauthoryear{Seegelke and W\"{u}hr}{Seegelke and
  W\"{u}hr}{2018}]%
        {Seegelke2018}
\bibfield{author}{\bibinfo{person}{Christian Seegelke} {and}
  \bibinfo{person}{Peter W\"{u}hr}.} \bibinfo{year}{2018}\natexlab{}.
\newblock \showarticletitle{Compatibility between object size and response side
  in grasping: the left hand prefers smaller objects, the right hand prefers
  larger objects}.
\newblock \bibinfo{journal}{\emph{{PeerJ}}}  \bibinfo{volume}{6}
  (\bibinfo{date}{Dec.} \bibinfo{year}{2018}), \bibinfo{pages}{e6026}.
\newblock
\urldef\tempurl%
\url{https://doi.org/10.7717/peerj.6026}
\showDOI{\tempurl}


\bibitem[\protect\citeauthoryear{Silver, Schrittwieser, Simonyan, Antonoglou,
  Huang, Guez, Hubert, Baker, Lai, Bolton, Chen, Lillicrap, Hui, Sifre, van~den
  Driessche, Graepel, and Hassabis}{Silver et~al\mbox{.}}{2017}]%
        {Silver17a}
\bibfield{author}{\bibinfo{person}{David Silver}, \bibinfo{person}{Julian
  Schrittwieser}, \bibinfo{person}{Karen Simonyan}, \bibinfo{person}{Ioannis
  Antonoglou}, \bibinfo{person}{Aja Huang}, \bibinfo{person}{Arthur Guez},
  \bibinfo{person}{Thomas Hubert}, \bibinfo{person}{Lucas Baker},
  \bibinfo{person}{Matthew Lai}, \bibinfo{person}{Adrian Bolton},
  \bibinfo{person}{Yutian Chen}, \bibinfo{person}{Timothy Lillicrap},
  \bibinfo{person}{Fan Hui}, \bibinfo{person}{Laurent Sifre},
  \bibinfo{person}{George van~den Driessche}, \bibinfo{person}{Thore Graepel},
  {and} \bibinfo{person}{Demis Hassabis}.} \bibinfo{year}{2017}\natexlab{}.
\newblock \showarticletitle{Mastering the game of Go without human knowledge}.
\newblock \bibinfo{journal}{\emph{Nature}} \bibinfo{volume}{550},
  \bibinfo{number}{7676} (\bibinfo{date}{Oct.} \bibinfo{year}{2017}),
  \bibinfo{pages}{354--359}.
\newblock
\urldef\tempurl%
\url{https://doi.org/10.1038/nature24270}
\showDOI{\tempurl}


\bibitem[\protect\citeauthoryear{Simpson}{Simpson}{1951}]%
        {Simpson51}
\bibfield{author}{\bibinfo{person}{E.~H. Simpson}.}
  \bibinfo{year}{1951}\natexlab{}.
\newblock \showarticletitle{The interpretation of interaction in contingency
  tables}.
\newblock \bibinfo{journal}{\emph{J. of the Royal Statistical Society, Series
  B}} \bibinfo{volume}{13}, \bibinfo{number}{2} (\bibinfo{year}{1951}),
  \bibinfo{pages}{238--241}.
\newblock


\bibitem[\protect\citeauthoryear{Sobel, Tenenbaum, and Gopnik}{Sobel
  et~al\mbox{.}}{2004}]%
        {journals/cogsci/SobelTG04}
\bibfield{author}{\bibinfo{person}{David~M. Sobel}, \bibinfo{person}{Joshua~B.
  Tenenbaum}, {and} \bibinfo{person}{Alison Gopnik}.}
  \bibinfo{year}{2004}\natexlab{}.
\newblock \showarticletitle{Children's causal inferences from indirect
  evidence: Backwards blocking and Bayesian reasoning in preschoolers}.
\newblock \bibinfo{journal}{\emph{Cognitive Science}} \bibinfo{volume}{28},
  \bibinfo{number}{3} (\bibinfo{year}{2004}), \bibinfo{pages}{303--333}.
\newblock


\bibitem[\protect\citeauthoryear{Uhde, Berberich, Ramirez-Amaro, and
  Cheng}{Uhde et~al\mbox{.}}{2020}]%
        {uhde_robot_iros2020}
\bibfield{author}{\bibinfo{person}{Constantin Uhde}, \bibinfo{person}{Nicolas
  Berberich}, \bibinfo{person}{Karinne Ramirez-Amaro}, {and}
  \bibinfo{person}{Gordon Cheng}.} \bibinfo{year}{2020}\natexlab{}.
\newblock \showarticletitle{The {Robot} as {Scientist}: {Using} {Mental}
  {Simulation} to {Test} {Causal} {Hypotheses} {Extracted} from {Human}
  {Activities} in {Virtual} {Reality}}. In \bibinfo{booktitle}{\emph{IROS
  2020}}. \bibinfo{address}{Las Vegas, USA}.
\newblock


\bibitem[\protect\citeauthoryear{Waismeyer, Meltzoff, and Gopnik}{Waismeyer
  et~al\mbox{.}}{2014}]%
        {Waismeyer2014}
\bibfield{author}{\bibinfo{person}{Anna Waismeyer}, \bibinfo{person}{Andrew~N.
  Meltzoff}, {and} \bibinfo{person}{Alison Gopnik}.}
  \bibinfo{year}{2014}\natexlab{}.
\newblock \showarticletitle{Causal learning from probabilistic events in
  24-month-olds: an action measure}.
\newblock \bibinfo{journal}{\emph{Developmental Science}} \bibinfo{volume}{18},
  \bibinfo{number}{1} (\bibinfo{date}{July} \bibinfo{year}{2014}),
  \bibinfo{pages}{175--182}.
\newblock
\urldef\tempurl%
\url{https://doi.org/10.1111/desc.12208}
\showDOI{\tempurl}


\bibitem[\protect\citeauthoryear{Walker, Lombrozo, Williams, Rafferty, and
  Gopnik}{Walker et~al\mbox{.}}{2016}]%
        {Walker2016}
\bibfield{author}{\bibinfo{person}{Caren~M. Walker}, \bibinfo{person}{Tania
  Lombrozo}, \bibinfo{person}{Joseph~J. Williams}, \bibinfo{person}{Anna~N.
  Rafferty}, {and} \bibinfo{person}{Alison Gopnik}.}
  \bibinfo{year}{2016}\natexlab{}.
\newblock \showarticletitle{Explaining Constrains Causal Learning in
  Childhood}.
\newblock \bibinfo{journal}{\emph{Child Development}} \bibinfo{volume}{88},
  \bibinfo{number}{1} (\bibinfo{date}{July} \bibinfo{year}{2016}),
  \bibinfo{pages}{229--246}.
\newblock
\urldef\tempurl%
\url{https://doi.org/10.1111/cdev.12590}
\showDOI{\tempurl}


\bibitem[\protect\citeauthoryear{W\"{u}hr and Seegelke}{W\"{u}hr and
  Seegelke}{2018}]%
        {Whr2018}
\bibfield{author}{\bibinfo{person}{Peter W\"{u}hr} {and}
  \bibinfo{person}{Christian Seegelke}.} \bibinfo{year}{2018}\natexlab{}.
\newblock \showarticletitle{Compatibility between Physical Stimulus Size and
  Left-right Responses: Small is Left and Large is Right}.
\newblock \bibinfo{journal}{\emph{Journal of Cognition}} \bibinfo{volume}{1},
  \bibinfo{number}{1} (\bibinfo{date}{Feb.} \bibinfo{year}{2018}),
  \bibinfo{pages}{17}.
\newblock
\urldef\tempurl%
\url{https://doi.org/10.5334/joc.19}
\showDOI{\tempurl}


\end{thebibliography}

%%%%%%%%%%%%%%%%%%%%%%%%%%%%%%%%%%%%%%%%%%%%%%%%%%%%%%%%%%%%%%%%%%%%%%%%

\end{document}